\documentclass{article}

\usepackage{PRIMEarxiv}

\usepackage[utf8]{inputenc} 
\usepackage[T1]{fontenc}    
\usepackage{hyperref}       
\usepackage{url}            
\usepackage{booktabs}       
\usepackage{amsfonts}       
\usepackage{nicefrac}       
\usepackage{microtype}      
\usepackage{lipsum}
\usepackage{fancyhdr}       
\usepackage{graphicx}       
\graphicspath{{media/}}     

\usepackage{pifont}
\usepackage{xcolor}

\usepackage{mfirstuc}

\usepackage[inline]{enumitem}

\usepackage[flushleft]{threeparttable}

\makeatletter
\newcommand{\expandCapFirst}[1]{%
    \expandafter\makefirstuc\expandafter{#1}}
\makeatother

\newcommand{\libname}{\emph{TemporAI}}

\newcommand{\ghHref}{\href{https://github.com/vanderschaarlab/temporai}{GitHub}}
\newcommand{\ghUrl}{\url{https://github.com/vanderschaarlab/temporai}}
\newcommand{\docsHref}{\href{https://temporai.readthedocs.io/}{project documentation}}
\newcommand{\docsUrl}{\url{https://temporai.readthedocs.io/}}

\newcommand{\hubHref}{\href{https://www.vanderschaar-lab.com/hub-for-healthcare/}{van der Schaar lab \emph{Hub for Healthcare}}}
\newcommand{\hubUrl}{\url{https://www.vanderschaar-lab.com/hub-for-healthcare/}}

\newcommand{\maintitle}{\libname{}: Facilitating Machine Learning Innovation in Time Domain Tasks for Medicine}

\newcommand{\topicIteMl}{causal inference}
\newcommand{\topicIteMed}{(individualized) treatment effect estimation}
\newcommand{\topicIteMedShort}{treatment effect estimation}
\newcommand{\topicPredMl}{prediction}
\newcommand{\topicPredMed}{forecasting}
\newcommand{\topicSurvMl}{time-to-event analysis}
\newcommand{\topicSurvMed}{survival analysis}
\newcommand{\topicInterpMl}{model interpretability}
\newcommand{\topicInterpMlShort}{interpretability}
\newcommand{\topicImputeMl}{data imputation}
\newcommand{\topicClusterMl}{clustering}
\newcommand{\topicClusterMed}{phenotyping}

\newcommand{\swClairvoyance}{\emph{clairvoyance}}
\newcommand{\swSeglearn}{\emph{seglearn}}
\newcommand{\swTsfresh}{\emph{tsfresh}}
\newcommand{\swSktime}{\emph{sktime}}
\newcommand{\swDarts}{\emph{darts}}
\newcommand{\swKats}{\emph{Kats}}

\newcommand{\swClairvoyanceCite}{\cite{jarrett2021clairvoyance}}
\newcommand{\swSeglearnCite}{\cite{burns2018seglearn}}
\newcommand{\swTsfreshCite}{\cite{christ2018time}}
\newcommand{\swSktimeCite}{\cite{loning2019sktime}}
\newcommand{\swDartsCite}{\cite{darts}}
\newcommand{\swKatsCite}{\cite{jiang_2021}}
\newcommand{\swSksurvCite}{\cite{sksurv}}
\newcommand{\swSksurvTbl}{\begin{tabular}[c]{@{}c@{}}\emph{scikit-}\\\emph{survival}\end{tabular}}
\newcommand{\swClairvoyanceTbl}{\begin{tabular}[c]{@{}c@{}}\emph{clair-}\\\emph{voyance}\end{tabular}}
\newcommand{\swSklearn}{\emph{scikit-learn}}
\newcommand{\swSklearnCite}{\cite{sklearn,sklearnApi}}

\newcommand{\dmT}{time series}
\newcommand{\dmS}{static}
\newcommand{\dmE}{event}

\newcommand{\refTab}{Table~}
\newcommand{\refFig}{Figure~}

\newcommand{\tblcheck}{\textcolor{black}{\ding{51}}}
\newcommand{\tblcross}{\textcolor{lightgray}{\ding{55}}}
\newcommand{\tblsome}{\textcolor{gray}{\textbullet}}

\newcommand{\modPred}{\texttt{prediction}}
\newcommand{\modSurv}{\texttt{survival}}
\newcommand{\modIte}{\texttt{treatment}}
\newcommand{\modCluster}{\texttt{clustering}}
\newcommand{\modInterp}{\texttt{interpretability}}
\newcommand{\modImpute}{\texttt{preprocessing.imputation}}

\hypersetup{
    citecolor=blue,
    colorlinks=true,
    linkcolor=blue,
    filecolor=cyan,      
    urlcolor=magenta,
}

\title{\maintitle
}

\author{
  Evgeny S. Saveliev \\
  DAMTP, University of Cambridge \\
  \texttt{es583@cam.ac.uk} \\
   \And
  Mihaela van der Schaar \\
  DAMTP, University of Cambridge \\
  The Alan Turing Institute \\
  \texttt{mv472@cam.ac.uk} \\
}

\begin{document}
\maketitle

\begin{abstract}
\libname~is an open source Python software library for machine learning (ML) tasks involving data with a time component, focused on medicine and healthcare use cases.  It supports data in \dmT, \dmS, and \dmE modalities and provides an interface for \topicPredMl, \topicIteMl, and \topicSurvMl, as well as common preprocessing utilities and \topicInterpMl~methods.  The library aims to facilitate innovation in the medical ML space by offering a standardized temporal setting toolkit for model development, prototyping and benchmarking, bridging the gaps in the ML research, healthcare professional, medical/pharmacological industry, and data science communities.  \libname~is available on \ghHref \footnote{\ghUrl} and we welcome community engagement through use, feedback, and code contributions.

\end{abstract}


\section{Time domain is crucial for ML in medicine}
\label{sec:intro}

Data with a time component\footnote{Depending on the context, referred to alternatively as: temporal, longitudinal, or time series data.} are ubiquitous in modern healthcare and medicine: from patient electronic health records (EHRs) \cite{moody2011physionet}, to data streams from Internet-of-Things (IoT) devices and consumer wearables \cite{dimitrov2016medical}, to large public health datasets \cite{dolley2018big}, naming just a few key growing areas.  In fact, since patient information is typically associated with a particular time point, the vast majority of healthcare data is temporal, and may be viewed as a time series.  Furthermore, availability of open access data in this field is also improving \cite{johnson2018mimic,saha2018md,sudlow2015uk}, attracting significant attention from the artificial intelligence (AI), machine learning (ML) and deep learning (DL) research, as well as the medical data science communities \cite{ching2018opportunities, piccialli2021survey, shickel2017deep}.  A such, it is evident that the temporal setting is becoming the cornerstone for ML in healthcare and medicine, with a significant potential for impact.

Numerous novel methods have been developed to tackle medically-relevant tasks in the time domain, such as: \topicPredMl~\cite{alaa2018hidden,bassiouni2022automated}, \topicIteMl~\cite{lim2018forecasting,bica2019estimating,zhang2022exploring}, \topicSurvMl~\cite{giunchiglia2018rnn,lee2019dynamic,wang2022survtrace}, \topicClusterMl~\cite{lee2020temporal,alqahtani2021deep}\footnote{In medical and other contexts, these tasks may also be referred to as, respectively: \topicPredMed, \topicIteMed, \topicSurvMed, \topicClusterMed. The descriptor ``temporal'' may be used to contrast with the static task setting.}, as well as \topicImputeMl~\cite{yoon2017multi,luo2018multivariate}, and \topicInterpMl~\cite{bento2021timeshap,crabbe2021explaining} methods, among others.  Yet currently a significant limitation exists in the lack of standardization of both data representation and model benchmarking \cite{ching2018opportunities, shickel2017deep}.  \libname~addresses these limitations as the first toolkit for development, prototyping and benchmarking of ML models on medically-relevant tasks with \dmT, \dmS, and \dmE data modalities.

\section{\libname{} library}
\label{sec:library}

\subsection{The ecosystem of temporal domain ML for medicine}
\label{sec:library-overview}
While a vibrant software ecosystem is developing for time series data science and applied machine learning, as exemplified by a range of libraries seen in \refTab \ref{tab:swtable}, few are primarily focused on the healthcare and medicine applications and problem settings.  \expandCapFirst{\swClairvoyance}~\cite{jarrett2021clairvoyance} developed by the authors previously was an attempt at filling this gap, but the library had significant drawbacks in
\begin{enumerate*}[label=(\roman*)]
  \item insufficient modularity in the code base,
  \item lack of robust testing and software engineering best practices,
  \item absence of continued support or proactive engagement with the community.
\end{enumerate*}
\libname~aims to address these issues, and to create a new ecosystem for temporal domain ML in medicine, aimed at -- and hoping to bridge the gap between -- the various communities in this field: ML researchers, research engineers, and data scientists, healthcare and pharmacology industry experts, as well as clinical professionals.

\begin{figure}[ht]
  \centering
  \includegraphics[width=0.95\textwidth]{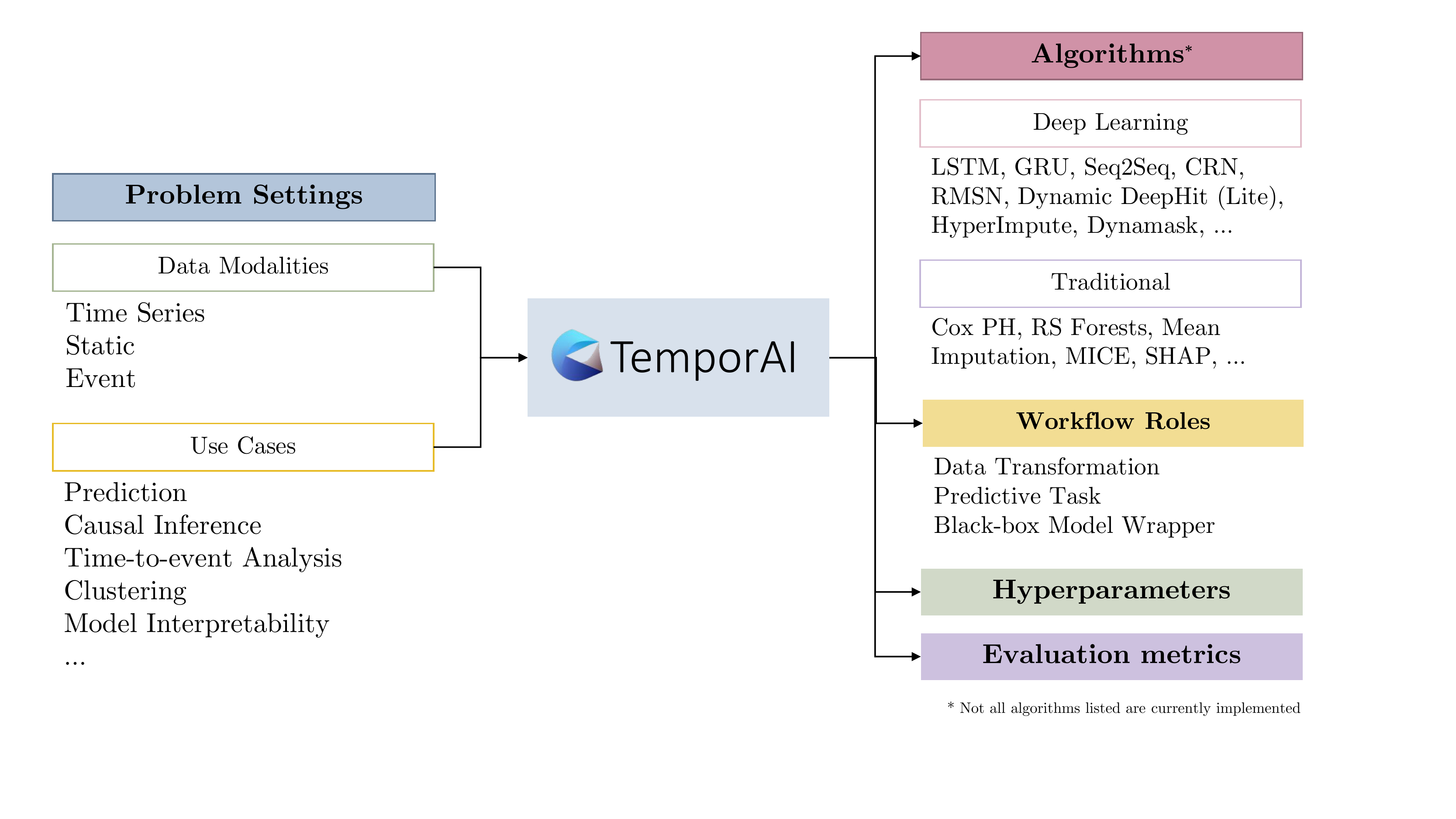}
  \caption{\libname~connects problem setting elements (data modality and use cases) to the appropriate algorithms, roles in the workflow, hyperparameters, and evaluation metrics.}
  \label{fig:figOverview}
\end{figure}

The design of \libname~maps the different data modalities and medically-relevant use cases to corresponding algorithm implementations, with their appropriate roles in the research workflow, and applicable evaluation metrics -- which we conceptually illustrate in \refFig \ref{fig:figOverview}.

\begin{table}
\begin{threeparttable}
 \caption{\libname~compared to a range of other time series frameworks}
 \label{tab:swtable}
  \centering
\begin{tabular}{@{}l|ccccccc|c@{}}
\toprule
                                 & \swSeglearn     & \swTsfresh     & \swSktime     & \swDarts     & \swKats     & \swSksurvTbl    & \swClairvoyanceTbl  & \libname  \\
                                 & \swSeglearnCite & \swTsfreshCite & \swSktimeCite & \swDartsCite & \swKatsCite & \swSksurvCite   & \swClairvoyanceCite &           \\ \midrule
\textbf{Primary focus:}          &                 &                &               &              &             &                 &                     &           \\
ML                               & \tblcross       & \tblcross      & \tblsome      & \tblsome     & \tblcheck   & \tblcross       & \tblcheck           & \tblcheck \\
Medical domain                   & \tblcross       & \tblcross      & \tblcross     & \tblcross    & \tblcross   & \tblsome        & \tblcheck           & \tblcheck \\ \midrule
\textbf{Software:}               &                 &                &               &              &             &                 &                     &           \\
Modular design                   & \tblcheck       & \tblcheck      & \tblcheck     & \tblcheck    & \tblcheck   & \tblcheck       & \tblcross           & \tblcheck \\
Unit and integration tests       & \tblsome        & \tblcheck      & \tblcheck     & \tblcheck    & \tblcheck   & \tblcheck       & \tblcross           & \tblcheck \\ \midrule
\textbf{Predictive tasks:}       &                 &                &               &              &             &                 &                     &           \\
\expandCapFirst{\topicPredMl}   & \tblcheck       & \tblcheck      & \tblcheck     & \tblcheck    & \tblcheck   & \tblcross       & \tblcheck           & \tblcheck \\
\expandCapFirst{\topicIteMl}    & \tblcross       & \tblcross      & \tblcross     & \tblcross    & \tblcross   & \tblcross       & \tblcheck           & \tblcheck \\
\expandCapFirst{\topicSurvMl}   & \tblcross       & \tblcross      & \tblcross     & \tblcross    & \tblcross   & \tblcheck$^{*}$ & \tblcross           & \tblcheck \\ \midrule
\emph{Additional functionality:} &                 &                &               &              &             &                 &                     &           \\
Preprocessing                    & \tblcheck       & \tblcheck      & \tblcheck     & \tblcheck    & \tblcheck   & \tblsome        & \tblsome            & \tblcheck \\
\expandCapFirst{\topicImputeMl} & \tblcross       & \tblsome       & \tblcross     & \tblcheck    & \tblcross   & \tblcross       & \tblcheck           & \tblcheck \\
\expandCapFirst{\topicInterpMl} & \tblcross       & \tblcheck      & \tblcross     & \tblcross    & \tblcheck   & \tblcross       & \tblcross           & \tblcheck \\ \bottomrule
\end{tabular}
    \begin{tablenotes}
      \item $^{*}$ Only with static covariates.
      \item \textbf{Key:} \tblcross~no, \tblsome~partial, \tblcheck~yes
    \end{tablenotes}
  \end{threeparttable}
\end{table}

\newcommand{\numSample}{N}
\newcommand{\numFeat}{F}
\newcommand{\idxSample}{i}
\newcommand{\idxFeat}{f}
\newcommand{\sFlag}{s}
\newcommand{\tFlag}{t}
\newcommand{\eFlag}{e}
\newcommand{\val}{x}
\newcommand{\idxFeatS}{\idxFeat_\sFlag}
\newcommand{\numFeatS}{\numFeat_\sFlag}
\newcommand{\valS}{\val_\sFlag}
\newcommand{\timeT}{t}
\newcommand{\idxFeatT}{\idxFeat_\tFlag}
\newcommand{\numFeatT}{\numFeat_\tFlag}
\newcommand{\idxTimeT}{\timeT_{\idxSample\idxFeat}}
\newcommand{\idxTimeTZero}{\timeT_{\idxSample\idxFeat}^{0}}
\newcommand{\idxTimeTMax}{\timeT_{\idxSample\idxFeat}^{max}}
\newcommand{\valT}{\val_\tFlag}
\newcommand{\timeE}{e}
\newcommand{\idxFeatE}{\idxFeat_\eFlag}
\newcommand{\numFeatE}{\numFeat_\eFlag}
\newcommand{\idxTimeE}{\timeE_{\idxSample\idxFeat}}
\newcommand{\valE}{\val_\eFlag}

\subsection{Data modalities}
\label{sec:library:data}
\libname~currently handles three data modalities: \emph{\dmT}, \emph{\dmS} and \emph{\dmE}:
\begin{enumerate}
    \item \emph{\expandCapFirst{\dmS}}: data comprised of features not associated with a time point. Each data point is described by $(\idxSample, \idxFeatS, \valS)$ where $\idxSample \in [\numSample]$ is the sample index, $\idxFeatS \in [\numFeatS]$ is the static feature index, and $\valS$ is the data point value. Patient blood type and year of birth are examples of static data modality features of a patient.
    
    \item \emph{\expandCapFirst{\dmT}}: data comprised of features, each of which is a sequence of values at certain time points. Each data point is described by $(\idxSample, \idxFeatT, \idxTimeT, \valT)$ where $\idxSample \in [\numSample]$ is the sample index, $\idxFeatT \in [\numFeatT]$ is the time series feature index, $\idxTimeT \in \{ \idxTimeTZero , \ldots , \idxTimeTMax \}$ is the time index of the data point (for sample $\idxSample$ and feature $\idxFeatT$), and $\valT$ is the data point value. Note that this representation is very flexible, and it means that:
    \begin{enumerate*}[label=(\roman*)]
      \item feature sequences may differ in length,
      \item time indexes need not be regular,
      \item and -- in the broadest case -- features of the same sample may not be aligned on their time indexes. A patient's systolic blood pressure measurements and white blood cell counts are examples of time series data modality features.
    \end{enumerate*}
    
    \item \emph{\expandCapFirst{\dmE}}: data comprised of features, each of which is represented by a time index and a value. Each data point is described by $(\idxSample, \idxFeatE, \idxTimeE, \valE)$ where $\idxSample \in [\numSample]$ is the sample index, $\idxFeatE \in [\numFeatE]$ is the event feature index, $\idxTimeE$ is the time index of the event (for sample $\idxSample$ and feature $\idxFeatE$), and $\valE$ is the data point value. Patient death or a myocardial infarction are examples of the event modality.
\end{enumerate}

The following data point value types are currently supported: continuous $\val \in \mathbb{R}$, integer $\val \in \mathbb{Z}$, and categorical $\val \in \{ c_1, \ldots , c_n \}$. A sentinel value $*$ is allowed, representing a missing data point (and in the context of the event modality, this naturally represents censoring).

\libname~allows for the capacity to support different data containers (e.g. \texttt{pandas.DataFrame} or \texttt{numpy.ndarray}), and different ways of representing the data modalities within these.

Finally, \libname~distinguishes three roles any set of data features may fulfil: \emph{covariates} \texttt{X}, \emph{targets} \texttt{Y}, and \emph{treatments} (interventions) \texttt{A}.

\subsection{Workflow}
\label{sec:library:workflow}

\begin{figure}[ht]
  \centering
  \includegraphics[width=0.95\textwidth]{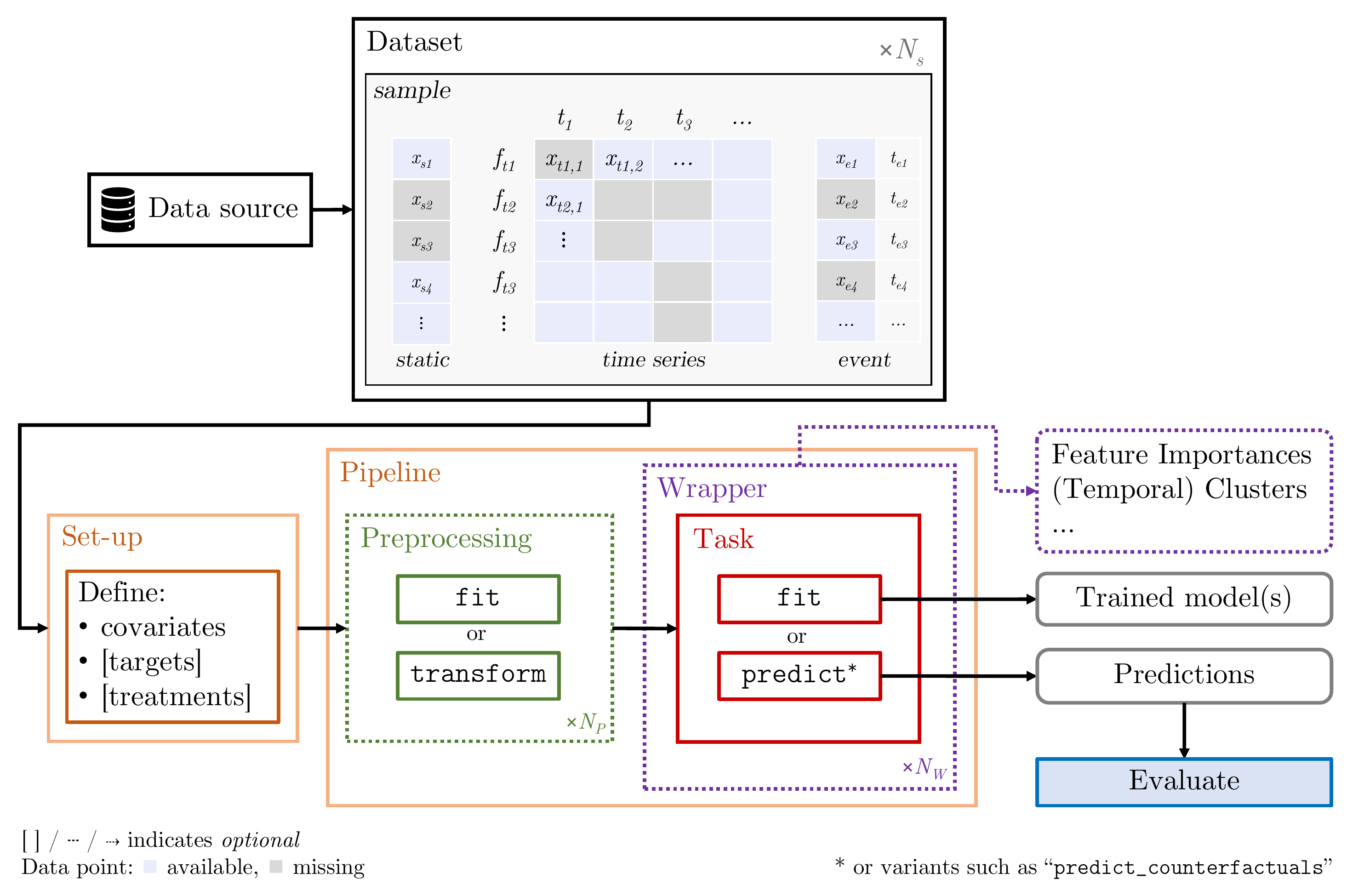}
  \caption{\libname~dataset and a typical workflow illustrated as a block diagram.}
  \label{fig:figPipe}
\end{figure}

The \libname~workflow builds on the familiar \texttt{fit/transform/predict} API of \swSklearn~\swSklearnCite. Additional methods are introduced when necessary to cover the relevant use cases, e.g. \texttt{predict\_counterfactuals} for \topicIteMl.  \refFig \ref{fig:figPipe} demonstrates a typical workflow. It is worth highlighting that \libname~introduces a novel role in the workflow: that of a ``wrapper'' model. This allows for implementation of algorithms which encapsulate other models, as may be the case in some \topicInterpMlShort~or \topicClusterMl~use cases. 

\subsection{Components}
\label{sec:library:comparison}

For an up-to-date information on the available library components (algorithms, data formats, etc.), please refer to \docsHref \footnote{\docsUrl}.

\subsection{Design principles}
\label{sec:library:design}

The library adheres to the software engineering design principles of \emph{modularity}, \emph{encapsulation}, and \emph{abstraction}, and emphasizes the ease of \emph{extensibility} -- as developing into a benchmarking environment in the machine learning for medicine temporal domain is one of its goals.

\section{Medical use cases}
\label{sec:use}

\libname~can be used to address a wide range of questions in the medical domains of acute and critical care, hospital medicine, transplantation, and preventative care, among others. 
Broadly, problems within the temporal setting involving predicting or forecasting, risk scoring (potentially with competing risks), individualized treatment effects, and patient phenotyping are in the scope of this tool. We maintain a reference of healthcare questions categorized by their medical domain, problem type, canonical formalism, and the most applicable software package at \hubHref \footnote{\hubUrl}, which we encourage the reader to brows.  This resource is being updated as we gather more feedback and guidance from medical experts.  A discussion of illustrative examples corresponding to the major areas of \libname~follows.

\subsection{\expandCapFirst{\topicPredMed}}
\label{sec:use:pred}

The \topicPredMed~(\topicPredMl) use case is well-exemplified by the intensive care setting. A typical problem of interest may be ``How can machine learning improve early warning scores using all the data clinicians have access to?'' The approach would be to utilize the available \dmS, \dmE, and \dmT~data modalities of a patient to predict either a static or a time series outcome which denotes the risk variable -- \libname~\modPred~module provides the tooling needed for training, inference, and evaluation of models in such a setting.

\subsection{\expandCapFirst{\topicSurvMed}}
\label{sec:use:surv}

The \topicSurvMed~(\topicSurvMl) use case can be illustrated by an example in transplantation. One may wish to find out ``How can I identify those patients who will need transplantation at an earlier stage of their illness?'' or ``How can we use the most up to date data to adapt predictions (usually based on data from 10-15 years prior to the time the prediction is made) so that they reflect probable future outcomes?''  Here, making use of the available patient \dmS, \dmE, and \dmT~data modalities, the goal would be to obtain accurate risk scores or survival curves, to aid in answering such questions.  \libname~\modSurv~module provides the tooling necessary for this setting.

\subsection{\expandCapFirst{\topicIteMedShort}}
\label{sec:use:ite}

A question matching the \topicIteMed, or \topicIteMl, use case in the critical care domain may be ``Can we predict whether and when a patient on mechanical ventilation can be safely extubated if a particular course of antibiotics is given?''  Starting with the \dmS, \dmE, and \dmT~data modalities of a patient, the goal in this formalism is to predict reliable counterfactual target (static or time series) variables, given alternative treatment plan inputs -- such counterfactuals then act to aid in the decision-making process.  \libname~\modIte~module provides the algorithms for this use case.

\subsection{Making the most of data and models}
\label{sec:use:misc}

It is often the case that algorithms for a particular use case alone are insufficient for real life application of ML.  The data may require significant preprocessing such as missing value imputation, and, especially in the high stakes clinical setting, the user may wish to go beyond predictions and to gain additional insights from, e.g. understanding the most important contributing features, or the allocation of patients to clusters based on their likely outcomes.  With this in mind, \libname~provides tools for preprocessing (e.g. \modImpute~module), interpretability (\modInterp~module), and clustering (\modCluster~module)\footnote{Some of these modules are still under active development.} to build up a software ecosystem suitable for ML researchers, data scientists, and healthcare professionals alike.

\section*{Acknowledgments}
We thank Prof. Eoin McKinney (Cambridge Institute of Therapeutic Immunology and Infectious Disease) for many invaluable discussions of medical applications and of clinically-relevant problems. We also thank Nick Maxfield for his tireless work on designing the van der Schaar Lab \emph{Hub for Healthcare} and for his diligent curation of the topics therein.

\small{%

\bibliographystyle{unsrt}  
\bibliography{main}  

}

\end{document}